\documentclass[letterpaper, 10 pt, conference]{ieeeconf}  % Comment this line out if you need a4paper

\IEEEoverridecommandlockouts                              % This command is only needed if 
\usepackage{graphicx}
\graphicspath{figures/}
\usepackage[caption=false,font=footnotesize]{subfig}
\usepackage{epsfig} % for postscript graphics files
\usepackage[dvipsnames]{xcolor}
\usepackage{times}
\usepackage{booktabs}
\usepackage{multicol}
\usepackage[pagebackref=false,
bookmarks=true,bookmarksopen=true,
bookmarksnumbered=true,
breaklinks=true,
colorlinks=true,
anchorcolor=blue,citecolor=black,
urlcolor=blue,linkcolor=blue,filecolor=blue,
menucolor=blue,
]{hyperref}
\usepackage{amsmath,bbm,amssymb}
\usepackage{algorithm}
\usepackage{algpseudocode}
\usepackage{xcolor}
\usepackage{soul}
\newcommand{\mb}[1]{{\mathbf{#1}}}
\newcommand{\mc}[1]{{\mathcal{#1}}}
\newcommand{\bs}[1]{{\boldsymbol{#1}}}

\DeclareMathOperator*{\minimize}{minimize~}

\DeclareMathOperator{\Tr}{Tr}

\title{\LARGE \bf
Adaptive and Risk-Aware Target Tracking with \\Heterogeneous Robot Teams
}

\author{Siddharth Mayya$^{1}$, Ragesh K. Ramachandran$^{2}$, Lifeng Zhou$^{1}$, Gaurav S. Sukhatme$^{2,3}$, Vijay Kumar$^{1}$ % <-this % stops a space
\thanks{This research was sponsored by the Army Research Lab through ARL DCIST CRA W911NF-17-2-0181.}% <-this % stops a space
\thanks{$^{1}$S. Mayya, L. Zhou and V. Kumar are with the GRASP Laboratory, University of Pennsylvania, Philadelphia, PA, USA {\tt\small\{mayya, lfzhou kumar\}@seas.upenn.edu.}}%
\thanks{$^{2}$R.K. Ramachandran and G.S. Sukhatme are with the Department of Computer   Science, University of Southern California (USC), Los Angeles, CA, USA {\tt\small\{rageshku, gaurav\}@usc.edu.}}%
\thanks{$^{3}$G.S. Sukhatme holds concurrent appointments as a Professor at USC and as an Amazon Scholar. This paper describes work performed at USC and is not associated with Amazon.}}%

\begin{document}

\maketitle
\thispagestyle{empty}
\pagestyle{empty}

%%%%%%%%%%%%%%%%%%%%%%%%%%%%%%%%%%%%%%%%%%%%%%%%%%%%%%%%%%%%%%%%%%%%%%%%%%%%%%%%
\begin{abstract}
We consider a scenario where a team of robots with heterogeneous sensors must track a set of hostile targets which induce sensory failures on the robots. In particular, the likelihood of failures depends on the proximity between the targets and the robots. We propose a control framework that implicitly addresses the competing objectives of performance maximization and sensor preservation (which impacts the future performance of the team).  Our framework consists of a predictive component---which accounts for the risk of being detected by the target, and a reactive component---which maximizes the performance of the team regardless of the failures that have already occurred. Based on a measure of the abundance of sensors in the team, our framework can generate aggressive and risk-averse robot configurations to track the targets. Crucially, the heterogeneous sensing capabilities of the robots are explicitly considered in each step, allowing for a more expressive risk-performance trade-off. Simulated experiments with induced sensor failures demonstrate the efficacy of the proposed approach.
\end{abstract}

%%%%%%%%%%%%%%%%%%%%%%%%%%%%%%%%%%%%%%%%%%%%%%%%%%%%%%%%%%%%%%%%%%%%%%%%%%%%%%%%
\section{INTRODUCTION}

Operating in disruptive and failure-prone environments is a fundamental requirement for achieving real-world autonomy in robotics systems~\cite{thrun2002probabilistic}. These disruptions can either be adversarial---such as in defense and security applications~\cite{sless2014multi,renganathan2017spoof,sung2019multi}, or non-adversarial---due to the complex and noisy nature of the real world~\cite{khalastchi2019fault}. In such settings, the deployment of multi-robot teams to accomplish tasks adds a layer of flexibility within the system, in the form of increased reliability via redundancy and reconfigurability~\cite{cortes2017coordinated}. \par 

Consequently, a significant amount of research effort has been dedicated towards the design of multi-robot coordination algorithms which account for the presence of disturbances, e.g.~\cite{ulusoy2012robust,dias2004robust,luo2019minimum}. Broadly speaking, these approaches consist of two possible elements: a \emph{predictive} risk-aware piece which generates solutions based on a model of the disturbance (typically interpreted as \textit{robustness})~\cite{zhou1998essentials}, or a \emph{reactive} piece which adapts the underlying control algorithm to continue task execution regardless of the disruptions affecting the system (typically interpreted as \textit{resilience})~\cite{chamon2020resilient}. The importance of both these aspects depends on the mission objectives and can change during operations---while the latter prioritizes task performance, the former generates potentially conservative solutions to prevent future disruptions.\par 

%%%%%%%%%%%%%%%%%%%%%%%%%%%%%%%%
	\begin{figure}
		\centering
		\includegraphics[width=1.0\linewidth]{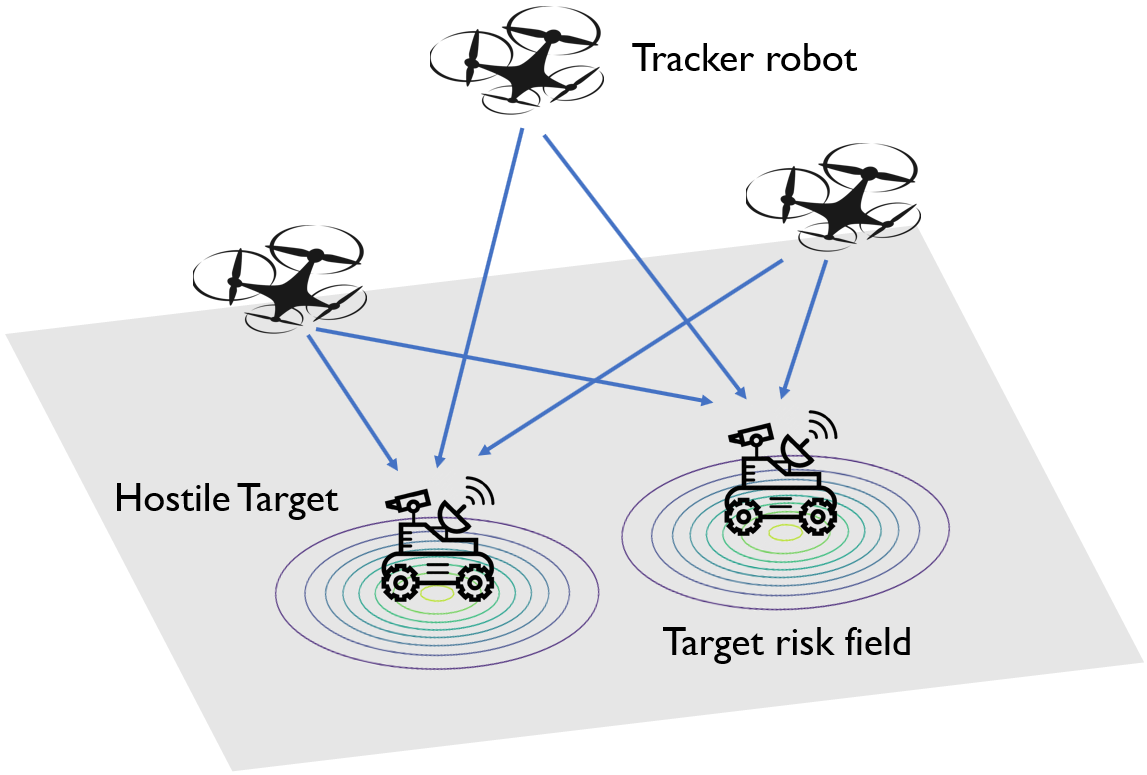}
		\caption{A depiction of the multi-robot target tracking scenario considered in this paper. A team of robots equipped with heterogeneous noisy sensors tracks a set of mobile targets. The targets are moving and can induce failures in the sensors of the robots based on the proximity between them. Therefore, the robots must position themselves close to the targets to generate accurate estimates, but must simultaneously account for the risk of failures---which impact the quality of future measurements.}
		\label{fig: illustration}       
	\end{figure}
	%%%%%%%%%%%%%%%%%%%%%%%%%%%%%%%%%%%

These considerations are highly relevant in the context of  \emph{heterogeneous} multi-robot teams, where the different capabilities of the robots can be leveraged to magnify the efficacy of the robot team~\cite{rizk2019cooperative}. For example, if only a single robot in the multi-robot team carries an essential sensor required for the success of the task, this robot's actions should be more risk-averse compared to other robots which carry non-essential sensors. A pertinent question then is: \emph{How should one systematically balance the objectives of risk-aversion and performance-focused adaptiveness in heterogeneous multi-robot teams?} \par 

In the context of a target tracking application, this paper presents a framework which synthesizes adaptive and risk-aware control inputs for a team of robots equipped with heterogeneous sensing capabilities. We consider a team of robots tracking mobile adversarial targets in a failure-prone environment. The robots are equipped with different types of noisy sensors, and are tasked with estimating the targets' states. Furthermore, the targets are moving and can induce failures in the sensors of the robots based on the proximity between them (see Fig.~\ref{fig: illustration}). In such a scenario, the robots must not only position themselves in order to achieve a trade-off between risk and tracking quality, they must also suitably adapt to failures that have rendered certain sensors dysfunctional. \par 
We endow the robot team with a Kalman filter~\cite{thrun2002probabilistic} %\cite{welch1995introduction}
which generates optimal estimates of the target locations along with a quantification of the uncertainty corresponding to each target's estimate. To embed risk-awareness in our framework, we introduce the notion of a \emph{safety-aware observability Gramian} (SOG), which weighs the quality of future observations made by the multi-robot team with the risk of failures associated with making those observations. By encoding the heterogeneous sensors available to the robots within the SOG, we enable our framework to position robots based on the contribution of their sensors to the overall tracking performance. \par 
Within our framework, the objectives of risk-aversion and tracking performance maximization are automatically traded off by considering the \emph{sensing margin}---defined as the excess number of operational sensors when compared to the minimum required for collective observability of the targets tracking process. This enables our algorithm to prioritize performance when the sensing margin is large and switch to risk-averse behaviors as the sensing margin of the team decreases. As we demonstrate, when compared to a framework with no risk-awareness, our algorithm preserves the sensors of the robots for longer, resulting in an extended operational time horizon. Moreover, our approach presents the mission designer with an explicit mechanism to tune this trade-off between performance and risk-aversion. \par 
The outline of the paper is as follows. Section~\ref{sec:lit_rev} places our work in the context of existing literature. Section~\ref{sec:ps} introduces the formulation for the target tracking scenario along with some mathematical notations. Section~\ref{sec:rrtt} presents the centralized optimization program whose solution generates adaptive and risk-aware configurations for the robot team. Section~\ref{sec:exp} presents the results of simulated experiments, and Section~\ref{sec:conclusion} concludes the paper.

\section{Literature Review}\label{sec:lit_rev}
Robust control and planning methods typically account for the risk of future failures and adversarial attacks occurring in the system when generating solutions~\cite{zhou1998essentials,yu2013tube}. Techniques such as $\mathcal{H}_{\infty}$ control~\cite{zhou1998essentials} and robust tube MPC~\cite{yu2013tube} have been specifically designed for ensuring operability under ``worst-case" disturbances or failures. Recently, similar traits have been investigated to develop robust coordination and planning algorithms in multi-robot systems. For instance, Zhou et al.~\cite{zhou2018resilient,zhou2020distributed} devised robust target tracking algorithms for robot teams to counter worst-case denial-of-service (DoS) failures or attacks that can make robots fail or compromise their tracking sensors. Their algorithms guarantee a provably  close-to-optimal team performance even though some robots or their sensors malfunction. Along this line, robust information collection algorithms~\cite{schlotfeldt2018resilient} were developed while accounting for worst-case robot failures or attacks when a team of robots are jointly planning paths to gather data in the environment. In contrast to these approaches which might generate conservative solutions,~\cite{zhou2018approximation} adopts a probabilistic approach, and presents a risk-aware coordination algorithm which addresses the issue of random sensor failures in the problem of sensor placement for environmental monitoring.\par
As disruptions are inevitable during the long duration operations of robot teams, mechanisms to maintain performance despite failures have been developed~\cite{ramachandran2020resilience,saulnier2017resilient,ramachandran2019resilience,song2020care}. To this end, Saulnier et al.~\cite{saulnier2017resilient} presented a resilient formation control algorithm that achieves the formation despite deceptive signals from non-cooperative team members. Similarly, Ramachandran et al.~\cite{ramachandran2019resilience} designed an algorithm to reconfigure the communication network of the robot team in order to compensate for resource failures. This algorithm was then utilized to adapt to failures of sensing resources in a multi-target tracking scenario~\cite{ramachandran2020resilience}. A resilient event-driven approach was devised by Song et al.~\cite{song2020care} to compensate for the robot failures when a team of robots are tasked to completely explore or cover an environment. Particularly, they designed a game-theoretic approach to trigger the reassignment of the functional robots, e.g., either continuing the coverage task in their pre-assigned workspace or being reassigned to the workspace of a failed robot.\par  

In contrast to these works which focus on either risk-awareness or adaptiveness, we present an optimization framework that integrates \emph{both} predictive and reactive control paradigms in the context of multi-robot multi-target tracking with heterogeneous sensing resources. In particular, our framework systemically exhibits a tradeoff between the performance maximization and the risk-aversion based on the abundance of resources available within the team.

\section{Problem Setup} \label{sec:ps}
This section formally introduces the heterogeneous sensing model of the multi-robot team and sets up the target tracking problem. Following this, we introduce the target-induced sensor failure models, which will be used in Section~\ref{sec:rrtt} to imbue risk-awareness into the proposed controller.
\subsection{Notation}
In this paper, capital or small letters, bold small letters and bold capital letters  represent scalars, vectors and matrices, respectively. Calligraphic symbols are used to denote sets. For any positive integer $z \in \mathbb{Z}^+$, $[z]$ denotes the set $\{1,2, \cdots, z\}$. $\|\cdot\|_p$ denotes the $p$-norm and the induced $p$-norm for vectors and matrices respectively. We drop the subscript on the norm when referring to the 2-norm. Additionally, $\|\mathbf{M}\|_F \triangleq \sqrt{trace(\mathbf{M}^T\mathbf{M})}$, outputs the Frobenius norm of a matrix $\mathbf{M}$.  $\mathbf{1}^{m_1}$ and $\mathbf{1}^{m_1 \times m_2}$ denotes the vector and matrix of ones of appropriate dimension, respectively. $\mb{I}_m$ denotes the identify matrix of size $m\times m$. The operator $|\cdot|$ gives the length of a vector applied on a vector and cardinality of a set when operated on a set. Given a vector $\mb{v}$, $Diag(\mb{v})$ represents a diagonal matrix with elements of $\mb{v}$ along its diagonal. Also, $\text{Tr}(\mb{M})$ results in the trace of $\mb{M}$. We use $\mb{M}(i,j)$ to denote the $i,j$ element of $\mb{M}$. Given a set of matrices $\{\mb{M}_1, \mb{M}_2, \ldots, \mb{M}_n\}$, $[\mb{M}_1, \mb{M}_2, \ldots, \mb{M}_n]$ and $[\mb{M}_1; \mb{M}_2; \ldots, \mb{M}_n]$ represent the matrices obtained through horizontal and vertical concatenation of the given matrices respectively. Since vectors and scalars are special cases of matrices, the definitions apply to them too.  Furthermore, $\mb{M}_1 \oplus \mb{M}_2 \oplus \ldots \oplus \mb{M}_n$ yields a block diagonal matrix constructed from the given set of matrices. Similarly, the operation $\mb{M}_1 \otimes \mb{M}_2$ denotes the Kronecker product between the matrices. 

\subsection{Heterogeneity Model}
Consider a team of $N$ robots engaged in the target tracking task, indexed by the set $\mc{R} :=[N]$%$\{1,\ldots,N\}$
. Let $\mb{x}_i \in \mc{X} \subseteq \mathbb{R}^p$ denote the state of robot $i$, and $\mb{u}_i \in \mc{U} \subseteq \mathbb{R}^q$ denote its control input, which steers the state according to the following control-affine dynamics:
\begin{equation}
	\dot{\mb{x}}_i = f(\mb{x}_i) + g(\mb{x}_i)\mb{u}_i, 
\end{equation}
where $f$ and $g$ are continuously differentiable vector fields. \par 
The robots are equipped with a set of heterogeneous sensors for tracking targets in the environment. Let $U$ denote the number of distinct types of sensors available within the team. Let $\bs{\Gamma} \in \{0,1\}^{N \times U}$ denote the \emph{sensor matrix} which describes the distribution of sensors over the robot team:
\begin{equation}
	\bs{\Gamma}_{ij} = \begin{cases}
		1,\quad\text{if robot $i$ possesses sensor $j$},\\
		0,\quad\text{otherwise}.
	\end{cases}
\end{equation}

\subsection{Target Tracking}
The robots are tasked with tracking a set of $M$ targets in the environment, which are indexed by the set $\mc{T} := [M]$. Let $\mb{e}_j \in \mc{X} \subseteq \mathbb{R}^p$ denote the state of target $j\in\mc{T}$, whose motion in the environment is described by the following linear dynamics:
\begin{equation}
	\dot{\mb{e}}_j = \mb{A}_{j}\mb{e}_j + \mathbf{B}_j\mb{u}^e_j + \mb{w}^e_j,
\end{equation}
where $\mb{A}_{j}$ and $\mb{B}_j$ are the process and control matrices, $\mb{w}^j_e$ is  the zero-mean independent Gaussian process noise with a covariance matrix  $\mb{Q}_j \in \mathbb{R}^{p \times p}$, and $\mb{u}^e_j \in \mc{U} \subseteq \mathbb{R}^q$ is the control input of target $j \in \mc{T}$. Stacking the target states and target control inputs into the vectors $\mb{e}$ and $\mb{u}^e$, respectively,  we get:
\begin{align}
    \label{eqn:target team dyn}
    \dot{\mb{e}} = \mb{A}\mb{e} + \mathbf{B}\mb{u}^e + \mb{w}^e,
\end{align}
where $\mb{A} \triangleq \mb{A}_1 \oplus \ldots \oplus \mb{A}_M$, $\mb{B} \triangleq \mb{B}_1 \oplus \ldots \oplus \mb{B}_M$  and $\mb{w}^e \triangleq [\mb{w}^e_1; \ldots; \mb{w}^e_M]$. As described before, we let $\mb{w}^e \sim \mc{N}(0,\mb{Q}),\mb{Q} \triangleq \mb{Q}_1 \oplus \mb{Q}_2 \oplus \ldots \oplus \mb{Q}_N$.
\par 
Within our framework, every robot $i$ makes an observation of every target $j$ according to the following linear observation model:
\begin{equation}
	\mb{y}_{ij} = \mb{H}_{ij} \mb{e}_j + \bs{\nu}_{ij},\quad i\in\mc{R},~j\in\mc{T},
\end{equation}
where $\mb{y}_{ij}$ is the measurement, $\bs{\nu}_{ij}$ is the measurement noise, and $\mb{H}_{ij}$ is the  measurement matrix corresponding to robot $i$'s measurements of target $j$ and will depend on the sensor suite available to robot $i$. \par 
Let $\bs{\gamma}_i$ denote the vector containing the ordered column indices of the sensor matrix corresponding to robot $i$,
	\begin{equation}
		\bs{\gamma}_i = \left[j\in[U]~\mid~\bs{\Gamma}_{ij} = 1 \right].
\end{equation}
In this paper, we assume the sensors to be linear, and associate a one-dimensional output matrix corresponding to each sensor available in the team, denoted by the set $\mc{S} := \{\mb{h}_1,\ldots,\mb{h}_{|\mc{S}|}\}$. 
Consequently, the measurement matrix for robot $i$ associated with the process of taking the measurements about the state of target $j$ can be constructed simply using the set $\mc{S}$ and $\bs{\gamma}_i$,
	\begin{equation}
		\mathbf{H}_{ij} = [\mb{h}_{\bs{\gamma}_i(1)},\ldots,\mb{h}_{\bs{\gamma}_i(|\bs{\gamma}_i|)}]^T.
\end{equation}
We model the measurement noise $\bs{\nu}_{ij}\sim \mc{N}(0,\mb{R}_{ij})$ to be a zero-mean Gaussian process where covariance $\mb{R}_{ij}$ of the sensor noise is distance dependent, and exponentially increases with the distance between the robots and the targets,
\begin{align}\label{eq:sensor_cov}
        &\mb{R}^{-1}_{ij} = Diag([{R}^{-1}_{ij1}, {R}^{-1}_{ij2}, \ldots, {R}^{-1}_{ij|\bs{\gamma}_i|}])
\end{align}
with
\begin{align}
		&\mb{R}^{-1}_{ijl} = w_l\exp{\left(-\lambda_l\|\mathbb{P}(\mb{x}_i) - \mathbb{P}(\mb{e}_j)\|\right)},~\forall l \in [|\bs{\gamma}_i|],
\end{align}
where $\lambda_l$ and $w_l$ determine the noise characteristics of each sensor $l$ in the sensor suite of robot $i$, and $\mathbb{P}$ is a projector operator that maps the state of the robot to it's position. While this paper chooses an exponential decay model to capture the sensor performance characteristics, any other continuously differentiable degradation function can be utilized instead. \par 
In ensemble form, the measurements of all the targets by robot $i$ can be written as:
\begin{equation}
	\mb{y}_i = \mb{{H}}_i \mb{e} + \bs{\nu}_i, ~\forall i\in\mc{R},
\end{equation}
where ${\mb{H}}_i = I_M \otimes \mathbf{H}_{ij}$, and $\bs{\nu}_i = [\bs{\nu}_{i1},\ldots,\bs{\nu}_{iM}]^T$. Here $I_M$ denotes the identity matrix of size $M$. The measurement equation for the team can be written as:
\begin{equation}
	\mb{y} = \mb{{H}} \mb{e} + \bs{\nu},
	\label{eq:team_measurements}
\end{equation}
where $\mb{y} = [\mb{y}_1, \mb{y}_2;\ldots; \mb{y}_N]$,  $\mb{{H}} = [\mb{{H}}_1; \mb{{H}}_2; \ldots; \mb{{H}}_N ]$ and $\bs{\nu} = [\bs{\nu}_1; \bs{\nu}_2;\ldots; \bs{\nu}_N]$. The team-wise measurement noise can be written as $\bs{\nu} \sim \mc{N}(0,\mb{R})$ where $\mb{R} = \mb{R}_{11}\oplus\mb{R}_{12}\oplus\cdots\oplus\mb{R}_{NM}$. Thus, given a sensor matrix $\boldsymbol{\Gamma}$, one can construct the corresponding robot output matrix and the team output matrix. These sensing models will be utilized in Section~\ref{sec:rrtt} to compute the estimates of the target positions and regulate the tracking performance of the robots. 

\subsection{Target-induced Risk Model}
\label{subsec: risk of fail}
As discussed in the introduction, each target can detect and induce failures within the robot team. Towards this end, let $\phi_{j}: \mc{X} \rightarrow [0, 1]$ represent the \emph{target risk field} which models the risk of a robot being detected by target $j \in \mc{T}$. Formally, $\phi_j(\mb{x}_i)$ represents the probability of robot $i$ with state $\mb{x}_i$ being detected by target $j$. In addition, we define the \textit{immunity field} $\pi_j : \mc{X} \rightarrow \mathbb{R}^+$ associated with $\phi_j(\mb{x}_i)$ as
\begin{align}
    \label{eqn: immunity field}
    \pi_j(\mb{x}_i) = -\log (\phi_j(\mb{x}_i)).
\end{align}
Note that, $\pi_j$ increases with the decrease in $\phi_j$. Thus, the immunity field $\pi_j$ encodes the safety of a robot in the vicinity of target $j$. Intuitively, it quantifies the ability of robot $i$ to evade detection from target $j$ while residing at $\mb{x}_i$. Depending on the context, we refer to this model as either the risk model or the safety model. Furthermore, we assume that the functional form of the target risk fields are known to the robot team (see Section~\ref{sec:exp} for particular choices of these functions). The next section formulates the optimization program to track the targets in an adaptive and risk-aware manner.

\section{Adaptive and Risk-Aware Target Tracking} \label{sec:rrtt}
Given the measurements of the targets as well as the risk model, this section first introduces the Kalman filter equations which will be used to measure the performance of the team. Subsequently, we introduce a safety-aware observability framework, which is then used to formulate the optimization program for generating the positions of the robots.
\subsection{Generating Target Estimates}
As discussed above, the primary objective of the robots is to generate accurate estimates of the positions of the targets. We achieve this using a Kalman filter which generates an optimal estimate for the states of all targets $\hat{\mb{e}}$ along with a team-wise covariance matrix $\mb{P}$. Typically, the estimation by KF contains prediction and update steps. In the prediction step, we have, 
\begin{align}
& \hat{\mb{e}}_{-} = \mb{A}\hat{\mb{e}}_0, \\
& \mb{P}_{-} =  \mb{A} \mb{P}_0 {\mb{A}}^T + \mb{Q}, 
\end{align}
where $\hat{\mb{e}}_{-} $ and $\mb{P}_{-}$ are (a priori) estimated states of the targets and the (a priori) team-wise covariance matrix at the next step, respectively; $\hat{\mb{e}}_0$ and $\mb{P}_0$ are the estimated states of the targets and the team-wise covariance matrix at the current step, respectively; and $\mb{Q}$ is the process noise covariance corresponding to all targets. Once the robots' measurements $\mb{y}$ (Eq.~\ref{eq:team_measurements}) are available, we have the KF update step as, 
\begin{align} \label{eqn:kf}
& \hat{\mb{e}} = \hat{\mb{e}}_{-} + \mb{K}\widetilde{\mb{y}}, \\
& \mb{P} = (\mb{I} - \mb{K} \mb{{H}})\mb{P}_{-} ,\label{eqn:post_cov}
\end{align}
where $\hat{\mb{e}}$ and $\mb{P}$ are the (a posteriori) estimated states of the targets and the (a posteriori) team-wise covariance matrix at the next step; $\widetilde{\mb{y}} := \mb{y} - \mb{{H}}\hat{\mb{e}}_{-}$ is measurement pre-fit residual; $\mb{K}: = \mb{P}_{-}{\mb{{H}}}^T(\mb{{H}}\mb{P}_{-}{\mb{{H}}}^T + \mb{R})^{-1}$ is the Kalman gain with $\mb{R}$ denoting the covariance matrix of the robot team measurement noise $\bs{\nu}$, as described in~\eqref{eq:team_measurements}. 

Notably, the team-wise aposteriori covariance matrix $\mb{P}$ can be expressed as $\mb{P} = \mb{P}_1 \oplus \mb{P}_2 \oplus \ldots \oplus \mb{P}_M$ with $\mb{P}_{j}$ denoting the (a posteriori) covariance matrix corresponding to all robots' joint estimate of target $j\in \mathcal{T}$.

\subsection{Imparting Risk-aware Behaviors}
In order to encode risk-awareness into our framework, we leverage the observability of the system of moving targets. More specifically, the observability Gramian associated with the multi-robot team tracking the targets is defined as: $\mb{O} \triangleq  \sum_{k=0}^{T-1} (\mb{A}^T)^k\mb{{H}}^T\mb{{H}}(\mb{A})^k$, where $T\geq 0$ is a suitably chosen time horizon. The positive definiteness of $\mb{O}$ is a necessary and sufficient condition for the observability of a deterministic linear system \cite{hespanha2018linear}. Intuitively, $\mb{O}$ measures the ease of estimating the initial state of a deterministic linear system from a given sequence of measurements on the state of the system. A measure of the observability Gramian matrix (such as the trace or the determinant) is typically used to measure the ease of estimating the initial state of the linear system~\cite{tzoumas2018resilient}. \par
In our scenario, proximity between the robots and the targets will prove detrimental to the ability of the robots to sense the targets in the future, due to the increasing risk of target-induced sensor failures. To encode this, we introduce a weighted observability Gramian, expressed as:
\begin{align}
    \mb{O}_{\bs{\Pi}} \triangleq  \sum_{k=0}^{T-1} (\mb{A}^T)^k\mb{{H}}^T\bs{\Pi}\mb{{H}}(\mb{A})^k,
    \label{eq:safety_gamm}
\end{align}
where $\bs{\Pi}$ is a positive definite matrix quantifying the relative safety of the different sensor-equipped robots. As the detection models described in Section~\ref{subsec: risk of fail} are purely dependent on the states of the targets and the robots, we can define $\bs{\Pi}$ as,
\begin{align}
    \label{eqn: safe weight matrix}
    \bs{\Pi} = \oplus_{i = 1}^{N}\left(\oplus_{j = 1}^{M}~ I_{|\bs{\gamma}_i|} \otimes \pi_j(\mb{x}_i)\right)
\end{align}
where ${\pi}_j(\mb{x_i})$ is the immunity field with respect to robot $i$ and target $j$, as defined in~\eqref{eqn: immunity field}. Note that, the block diagonal matrices $I_{|\bs{\gamma}_i|} \otimes \pi_j(\mb{x}_i)$ which compose the matrix $\bs{\Pi}$ represent the inverse risk associated with target $j$'s detection of robot $i$. \par 
In other words, larger the diagonal entries of  $\mb{O}_{\bs{\Pi}}$,  larger is the ability of the robot team to obtain the targets' state measurements while evading detection by the targets. Thus, we refer to the matrix $\mb{O}_{\bs{\Pi}}$ as \textit{safety aware observability gramian} (SOG) matrix. In the optimization problem described in Section~\ref{sec:rrtt},  we bound the trace of $\mb{O}_{\bs{\Pi}}^{-1}$ from above to control the risk-averseness of the robot team.

\subsection{Optimization Problem}

We propose solving the following optimization program to obtain the states of the robots for target tracking. In the following, we will describe the various constraints and symbols constituting the problem description.

\begin{subequations}\label{eq:rtargettrackingopt}
	\begin{flalign}
		\text{\bf Adaptive and Risk-Aware Target Tracking} \tag{\ref{eq:rtargettrackingopt}}&&
		\label{eq:full_opt_target_tracking}
	\end{flalign}
	\begin{align}
		\begin{split}
			\minimize_{\substack{\mb{x},\boldsymbol{\delta}_1,\delta_2}} & k_1(\Delta, \boldsymbol{\delta}_1) + k_2(\Delta, \delta_2) 
		\end{split}\label{eq:nlp:a} \\
		% \|a\|_1 + k_1(\Delta, \boldsymbol{\delta}_1) + k_2(\Delta, \delta_2) 
		s.t. ~
		& \|\mb{x}_i - \mb{x}_{i0}\| \leq d_m, ~\forall i \in \mathcal{R},  \label{eq:nlp:b}\\
		& \|\mb{x}_i -\mb{x}_j\| \geq d_n, ~\forall i\neq j,~ \forall i,j \in \mathcal{R}, \label{eq:nlp:c}\\ 
		& \Tr(\mb{P}_j) \leq \bs{\rho}_{1,j} + {\delta}_{1,j}, ~\delta_{1,j} \geq 0, ~j\in\mathcal{T}, \label{eq:nlp:d}\\
		&\Tr(\mb{O_{\bs{\Pi}}}^{-1})  \leq \rho_2 + \delta_2, ~\delta_2 \geq 0, \label{eq:nlp:e}
	\end{align}
\end{subequations}
where $\mb{x}_{i0}$ is robot $i$'s state at the current time step, $\mb{x} := [\mb{x}_1^T, \mb{x}_2^T, \cdots, \mb{x}_N^T]^T$ is the optimization variable representing the ensemble state of all robots at the next time step, $d_m$ is the maximum distance each robot $i$ can move between two consecutive time steps, and $d_n$ is the minimum inter-robot distance for safety. Furthermore, $\bs{\rho}_{1} = [\bs{\rho}_{1,1}, \bs{\rho}_{1,2}, \cdots, \bs{\rho}_{1,M}]^T$  and $\rho_2$ denote the user-defined upper bounds on the tracking error (encoded by $\mb{P}$) and team-level risk (encoded by $\mb{O}_{\bs{\Pi}}^{-1}$), respectively. We would like our adaptive framework to continue operations even when sensor failures might make it impossible to achieve desired specifications. To this end, we introduce slack variables $\boldsymbol{\delta}_{1}$ and $\delta_2$ where $\boldsymbol{\delta}_1 := [\bs{\delta}_{1,1}, \bs{\delta}_{1,2}, \cdots, \bs{\delta}_{1,M}]^T$ introduces a separate performance slack variable for each target.\par

We now describe the various constraints introduced  in~\eqref{eq:full_opt_target_tracking}. Equation~\ref{eq:nlp:b} limits the distance that each robot can travel between two consecutive steps. Equation~\ref{eq:nlp:c} guarantees that the generated solutions maintain a minimum safety distance between the robots. Equation~\ref{eq:nlp:d} nominally aims to bound the trace of the posterior covariance  matrix $\mb{P}_j$~\eqref{eqn:post_cov} for each target $j$ by a constant $\bs{\rho}_{1,j}$. However, sensor failures might imply that meeting this performance criteria is not longer possible. To this end, the slack variable $\bs{\delta}_1$ is introduced to ensure that the framework adapts to the current operating conditions. Similarly, the risk in the team tracking performance is captured by the inverse SOG $\mb{O}^{-1}_{\bs{\Pi}}$ and it is upper bounded by the sum of $\rho_2$ and the slack variable $\delta_2$ to ensure a desired level of safety~\eqref{eq:nlp:e}. It should be noted that $\mb{P}$ is computed directly using the update equation~\eqref{eqn:post_cov} and $\mb{O}_{\bs{\Pi}}$ is computed using the estimated state of the targets $\hat{\mb{e}}$ computed in~\eqref{eq:safety_gamm}.

\subsection{Objective Function Design}
As described in the introduction, balancing the competing objectives of performance maximization and risk-aversion is a desirable feature of a resource-aware control framework. In this paper, we enable our framework to implicitly make this trade-off based on a quantification of the abundance of sensors available in the team. This is encoded within the cost functions $k_1$ and $k_2$ as described next. \par 
First, we define the resource abundance $\Delta$ as the difference between the current sensor matrix $\bs{\Gamma}$ and the minimal heterogeneous resources required for collective observability of the targets $\bs{\Gamma}_{\min}$,
\begin{equation}
	\Delta = \|\bs{\Gamma}\|_\mathcal{F} -\|\bs{\Gamma}_{\min}\|_\mathcal{F},
	\label{eq:resource_abundance}
\end{equation}
where $\bs{\Gamma}_{\min}$ is defined as, 
\begin{align}
\begin{split}
	\bs{\Gamma}_{\min} = & \arg\min_{\bs{\Gamma}} \|\bs{\Gamma}\|^2_{\mc{F}}  \\%\text{trace}((\mb{A}_e^T)^k\bs{\mc{H}}^T\bs{\mc{H}}(\mb{A}_e)^k) \\
	\label{eq:min_gamma}
	& s.t. ~\text{det}\left(\sum_{k=0}^{T-1}(\mb{A}^T)^k{\mb{H}}^T{\mb{H}}(\mb{A})^k\right) > 0. 
	\end{split}
\end{align}
Notably, \eqref{eq:min_gamma} computes the smallest set of sensors which will yield collective observability of the system.\par 

Generally speaking, with fewer heterogeneous resources (i.e., a lower $\Delta$), the robot team should lower its expectations on the tracking performance by choosing a larger $\|\boldsymbol{\delta}_1\|_1$ (in~\eqref{eq:nlp:d}) and be more risk-averse by choosing a smaller $\delta_2$ (in~\eqref{eq:nlp:e}). Conversely, with more heterogeneous resources, the robots should aim to achieve a higher tracking quality by choosing a smaller $\|\boldsymbol{\delta}_1\|_1$ and can be more risk-seeking by choosing a larger $\delta_2$. To this end, $k_1(\Delta, \boldsymbol{\delta}_1)$ belongs to a class of functions $f: [0, +\infty] \times  [0, +\infty] \to [0, +\infty]$ that is monotone increasing in both $\|\boldsymbol{\delta}_1\|_1$ and $\Delta$. While, $k_2(\Delta, \delta_2)$ belongs to a class of functions $f: [0, +\infty] \times  [0, +\infty] \to [0, +\infty]$ that is monotone increasing in $\delta_2$ and yet monotone decreasing in $\Delta$. Simple examples of $k_1(\Delta, \boldsymbol{\delta}_1)$ and $k_2(\Delta, \delta_2)$ can be
\begin{align}
	k_1(\Delta, \boldsymbol{\delta}_1) = w_1 \Delta \|\boldsymbol{\delta}_1\|_1, \\
	k_2(\Delta, \delta_2) = w_2 \frac{1}{\Delta} \delta_2^2,
	\label{eq:hexamples}
\end{align}
where $w_1, w_2$ are positive scalars. Notably, optimizing over such functions in the objective~\eqref{eq:nlp:a} enables the robots to make adaptive and risk-aware decisions given the abundance of the heterogeneous resources they have. While this paper utilizes a simple instantiation of the functions $k_1$ and $k_2$, more complex functions can be designed for achieving fine-tuned behaviors. 

\subsection{Computational Aspects}
\begin{figure}
		\centering
		%[width=0.9\linewidth,height=0.9\linewidth]
		\includegraphics[width=0.4\textwidth]{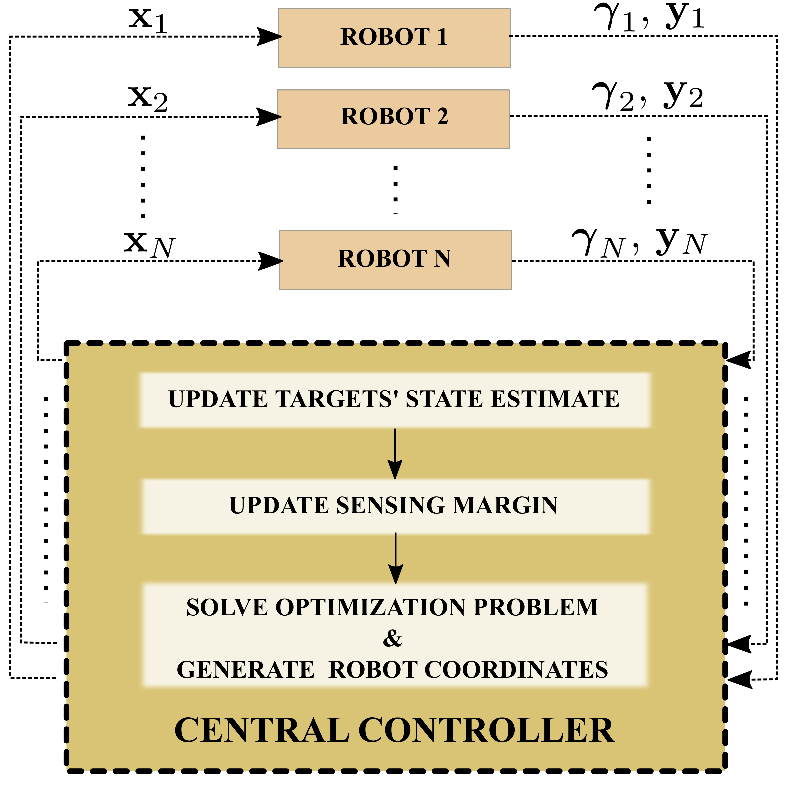}
		\caption{Operations of the proposed adaptive and risk-aware target tracking strategy. A central controller estimates the targets' state and sensing margin based on the information received from the robots. Using the targets' state estimate and the sensing margin, the central controller solves the optimization problem depicted in~\eqref{eq:full_opt_target_tracking} to generate new coordinates for the robots.} %\LZ{sorry, still, make the text as large as possible}}
		\vspace{0mm}
		\label{fig: schematic}       
	\end{figure}

The optimization problem presented in~\eqref{eq:full_opt_target_tracking} can be solved at discrete time intervals $t\in\mathbb{N}$ based on the updated estimates of the target states. Algorithm~\ref{alg:target_track} illustrates the operations of the target tracking framework. Step~\ref{step:kf} uses the Kalman filter to generate estimates of the target positions, $\hat{\mathbf{e}}$. Using this estimate, Step~\ref{step:opt} generates new positions for the robots according the tradeoffs between performance maximization and risk-aversion. We assume that the robots drive to the configuration generated by the optimization program within this step (ensured by ~\eqref{eq:nlp:a})~\cite{wang2017safety}. Following this, the robots evaluate which of their sensors have failed due to proximity with the targets, and update the sensing margin $\Delta$ in Step~\ref{step:delta}. See Fig.~\ref{fig: schematic} for a system diagram illustrating these operations.
\begin{algorithm}
	\caption{Adaptive and Risk-Aware Target Tracking}
	\label{alg:target_track}
	\begin{algorithmic}[1]
		\Require
		\Statex Robot team heterogeneity specifications: $\mb{\Gamma}, \mb{H}$
		\Statex Target and Team Specs: $N, M, \phi_j,w_l,\lambda_l$, $d_m, d_n$ 
		\Statex Parameters: $\bs{\rho}_1,\rho_2,w_1,w_2$
		\State Initialize: $t = 0$
		\While{true}
		\State Update target position estimate $\hat{\mb{e}}$\label{step:kf} \Comment \eqref{eqn:kf}
		\State Execute the adaptive risk-aware controller\label{step:opt}\Comment \eqref{eq:full_opt_target_tracking}
		\State Compile sensor failures and update $\boldsymbol{\Gamma}$
		\State Update sensing margin $\Delta$\label{step:delta}\Comment \eqref{eq:resource_abundance}
		\State $t = t + 1$
		\EndWhile
	\end{algorithmic}
\end{algorithm}

%%%%%%%%%%%%%%%%%%%%%%%%%%%%%%%%

\section{Simulated Experiments} \label{sec:exp}
In this section, we present a series of simulated experiments which demonstrate the ability of the proposed framework to track multiple targets in an adaptive and risk-aware manner using a team of robots with heterogeneous sensors. We simulate a team of ground robots operating in $\mathbb{R}^2$, with single integrator dynamics: $\dot{\mb{x}} = \mb{u}$, where $\mb{x}$ denotes the positions of the robots. The robots are equipped with three different types of sensors, whose reduced measurement matrices are given as:
\begin{align} \label{eq:sensors}
    h_1 = \begin{bmatrix} 1 & 0 \end{bmatrix},~h_2 = \begin{bmatrix} 0 & 1 \end{bmatrix},~h_3 = \begin{bmatrix} 1 & 1 \end{bmatrix}.
\end{align}
We enable the centralized controller (Fig.~\ref{fig: schematic}) to run a Kalman filter using the noisy measurements obtained from the robots and update the posteriori covariance matrix $\mb{P}$. \par 
The target risk field $\phi_j$ corresponding to each target $j\in\mc{T}$ is described by a Gaussian function centered around the target,
\begin{align}
    \phi_j(\mb{x}) = \frac{c_j}{2\pi|\Sigma_j|}\exp\left(-\frac{1}{2}(\mb{x} - \mb{e}_j)^T\Sigma_j(\mb{x}-\mb{e}_j)\right),
\end{align}
where $c_j$ determines the peak value of the field for target $j$, and $\Sigma_j$ determines the variance. The optimization problem given in~\eqref{eq:full_opt_target_tracking} was solved using SNOPT~\cite{snopt77} via pydrake~\cite{drake}.\par 
\subsection{Two robots, two targets} \label{subsec:exp1}
For the first set of experiments, we consider a team of two robots, whose sensor matrix at deployment time is specified as $\mb{\Gamma} = \mb{1}^{2 \times 3}$. The target risk field parameters are chosen  as $c_j = 3$ and $\Sigma_j = Diag([2.0, 2.0]),~\forall j\in\mc{T}$. Sensor failures on the robots are simulated by flipping a biased coin proportional to the magnitude of the target risk field at the robot's location. While the true state of the targets $\mb{e}$ is used to simulate the failures, the robots compute $\mb{O}_{\bs{\Pi}}$ using the estimated state of the targets $\hat{\mb{e}}$ as they do not have access to the true target states.\par
\begin{figure}[h]
\hspace*{-0.45cm}
	\centering
	\subfloat[][]{
		\includegraphics[trim={3.00cm 0cm 4.0cm 0cm},clip,width=0.24\textwidth]{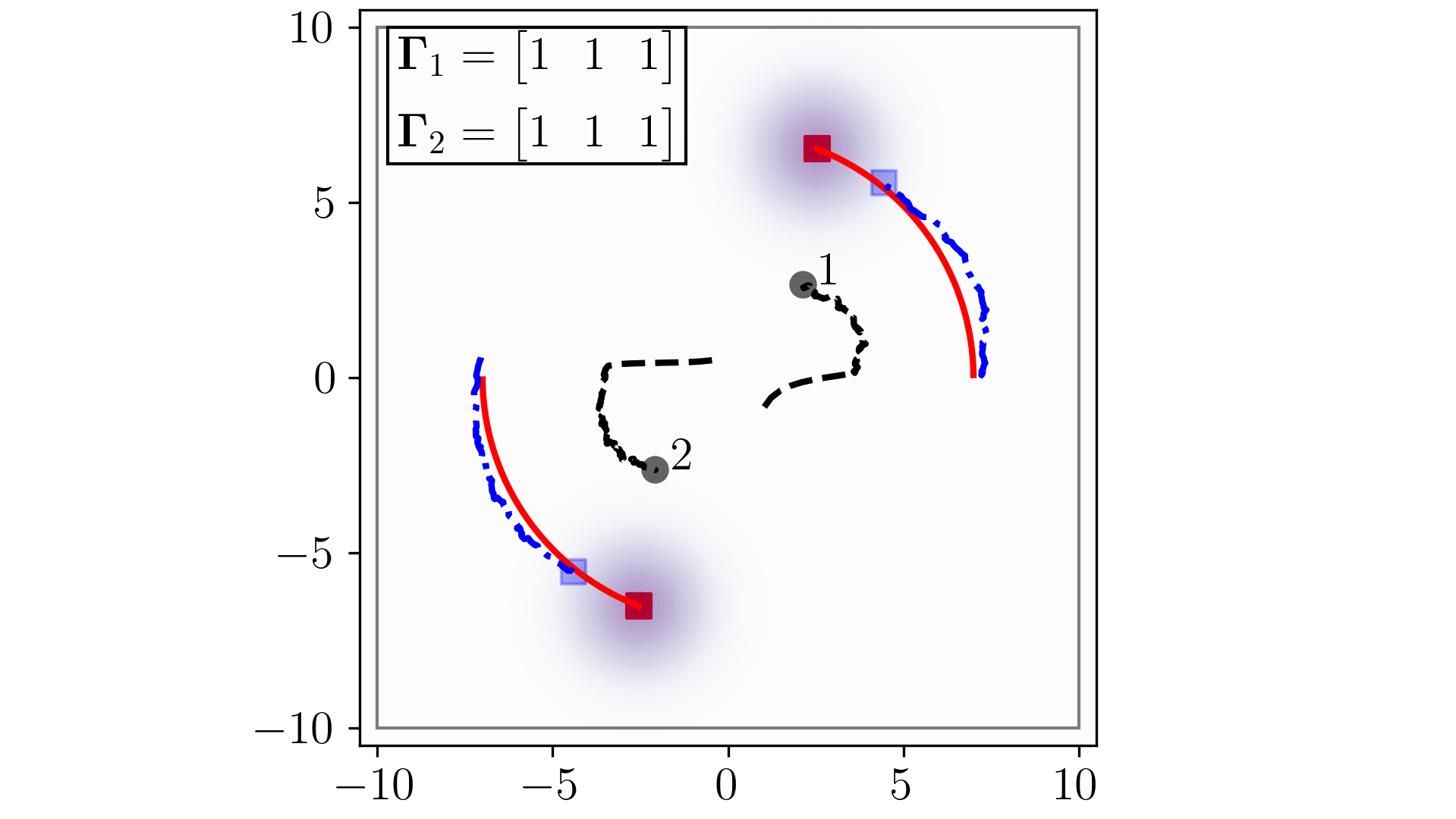}
		\label{fig:scenario_risk_aware}}
	\subfloat[][]{
		\includegraphics[trim={3.00cm 0cm 4.0cm 0cm},clip,width=0.24\textwidth]{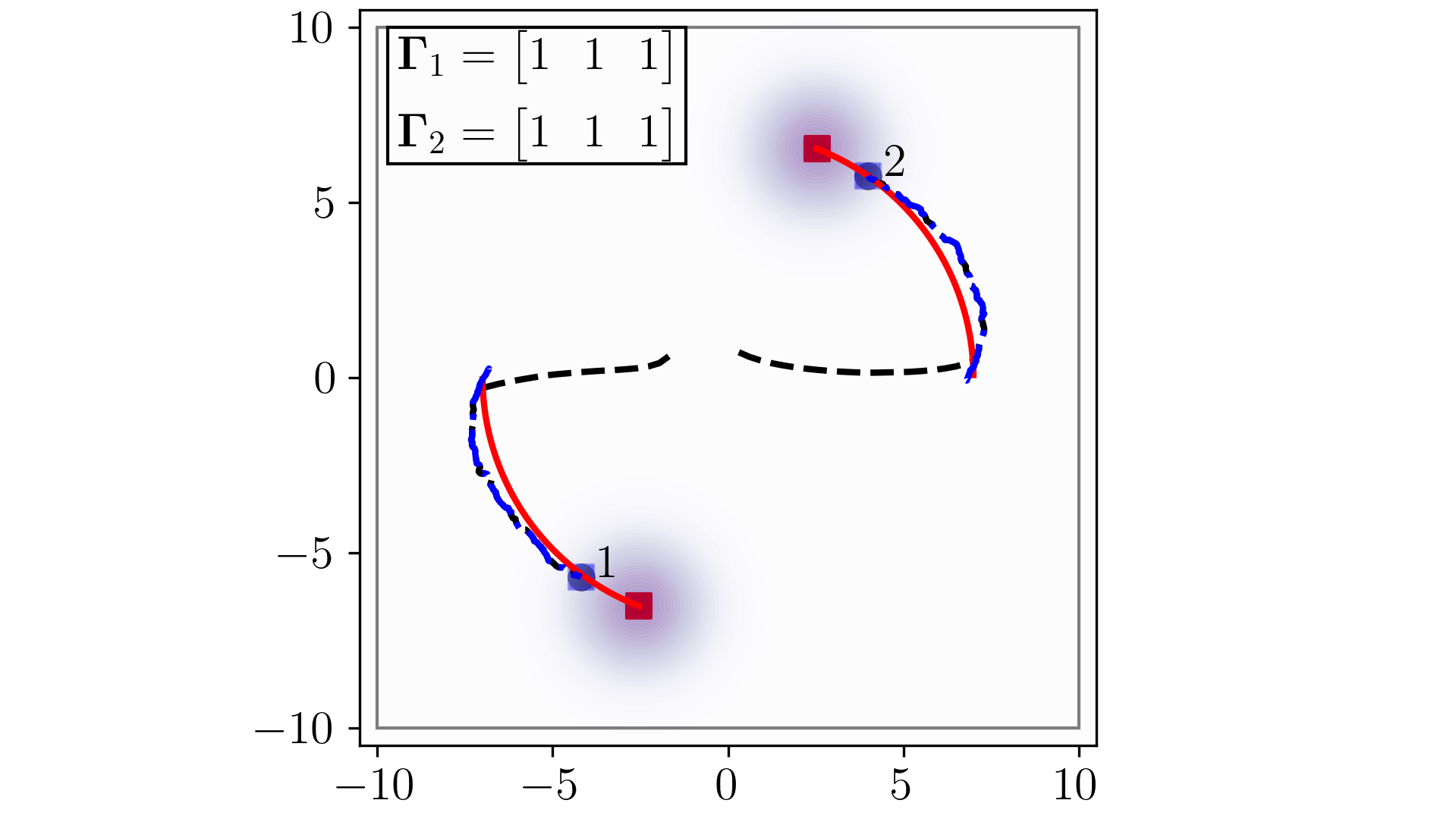}
		\label{fig:scenario_no_risk}}
	\caption{Motion trails of the robots (black dots), mobile targets (red squares), and estimated target locations (blue translucent squares) executing the adaptive \& risk-aware control framework presented in this paper. The target risk fields are illustrated by the shaded regions around the targets. \protect\subref{fig:scenario_risk_aware} illustrates how the risk-aware constraint~\eqref{eq:nlp:e} enables the robots to trade-off between current and future tracking performance by maintaining a distance from the targets. This can be constrasted with the behavior in~\protect\subref{fig:scenario_no_risk} where the risk-aware constraint has been removed. Consequently, the robots position themselves very close to the targets to optimize for tracking performance but at the risk of higher sensor failures rates which detrimentally affects their future performance.}
	\label{fig:scenarios}
\end{figure}
\begin{figure}[h]
\centering
	\includegraphics[width=0.44\textwidth]{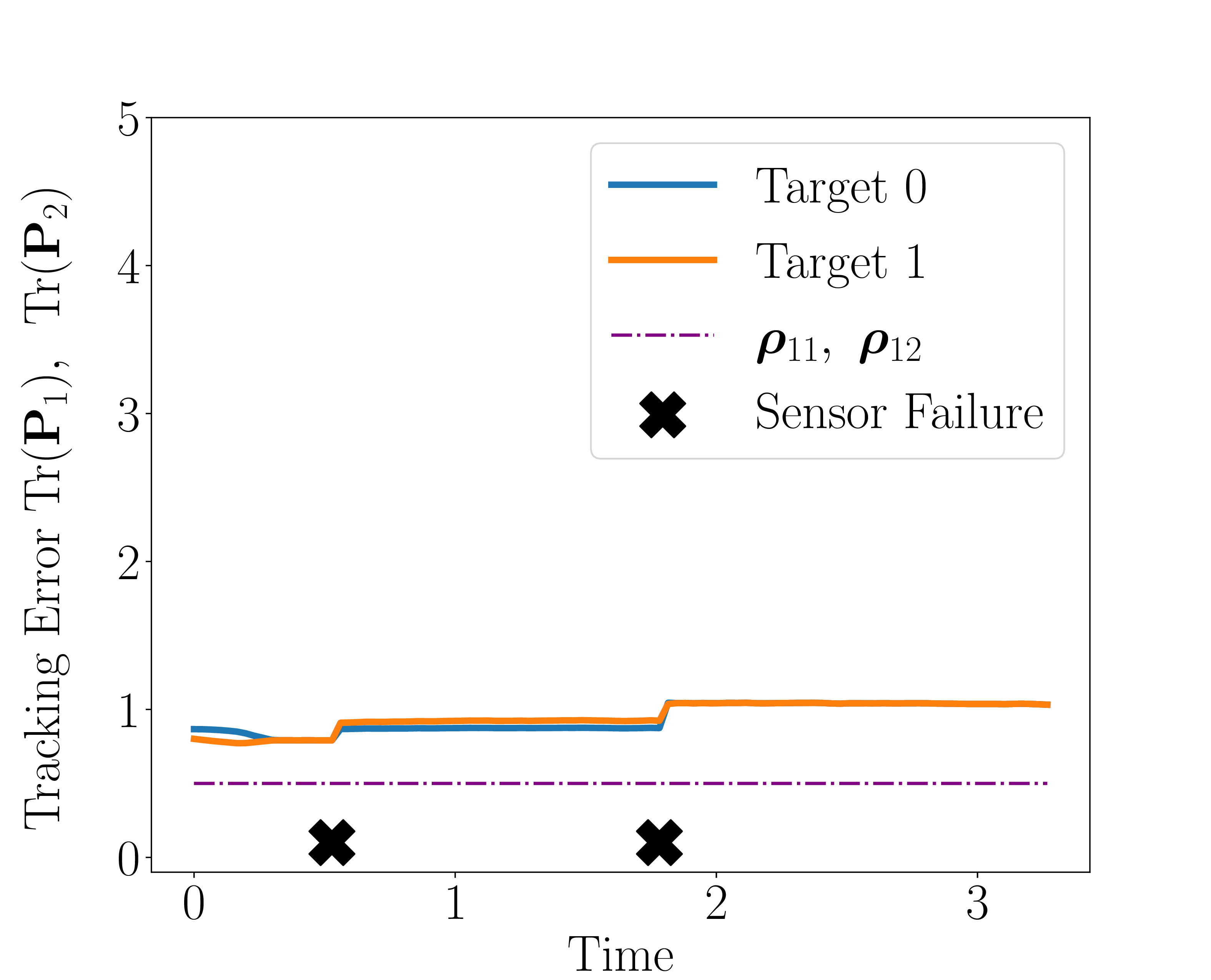}
	\caption{Tracking error for a team of two robots executing the adaptive \& risk-aware target tracking control presented in~\eqref{eq:rtargettrackingopt}. As seen, the robots do not achieve their tracking performance objectives $\bs{\rho}_1$ due to the safety Gramian constraint imposed in~\eqref{eq:nlp:e} as well as sensor failures induced by the targets. However, the adaptive nature of the performance optimization allows the team to continue tracking the targets while balancing the objectives of current performance and safe future operations.}
	\label{fig:tracking_perf}
\end{figure}

The parameters for the sensor covariance matrices are chosen as $w_l = 1.80, \lambda_l = 0.1, \forall l \in [|\mc{S}|]$, and the target dynamics are set to $\mb{A}_j = Diag([1, 1]), \mb{B}_j = Diag([1, 1]),~\forall j\in\mc{T}$. For the parameters $(d_m = 0.33,d_n = 2.0, \bs{\rho}_1 = [0.5,0.5], \rho_2 = 0.1, w_1 = 1 , w_2 = 100)$, Fig.~\ref{fig:scenario_risk_aware} illustrates the motion trails of the robots (black dots) relative to the true target positions (red squares) and the estimated target positions (blue translucent squares) as the simulation progresses. As seen, the robots approach the targets while accounting for the risk of detection and failures (depicted by the shaded purple regions).
 Figure~\ref{fig:tracking_perf} illustrates how the tracking error corresponding to each target---measured by $\Tr(\mb{P}_j)$---increases as failures affect the system. It should be noted that, despite not meeting the desired maximum tracking error $\bs{\rho}_1$, the team continues to perform the task, thanks to the slack variable $\boldsymbol{\delta}_1$ introduced in constraint~\eqref{eq:nlp:d}. \par 
\begin{figure}[h]
	\centering
	\subfloat[][]{
	\includegraphics[width=0.45\textwidth]{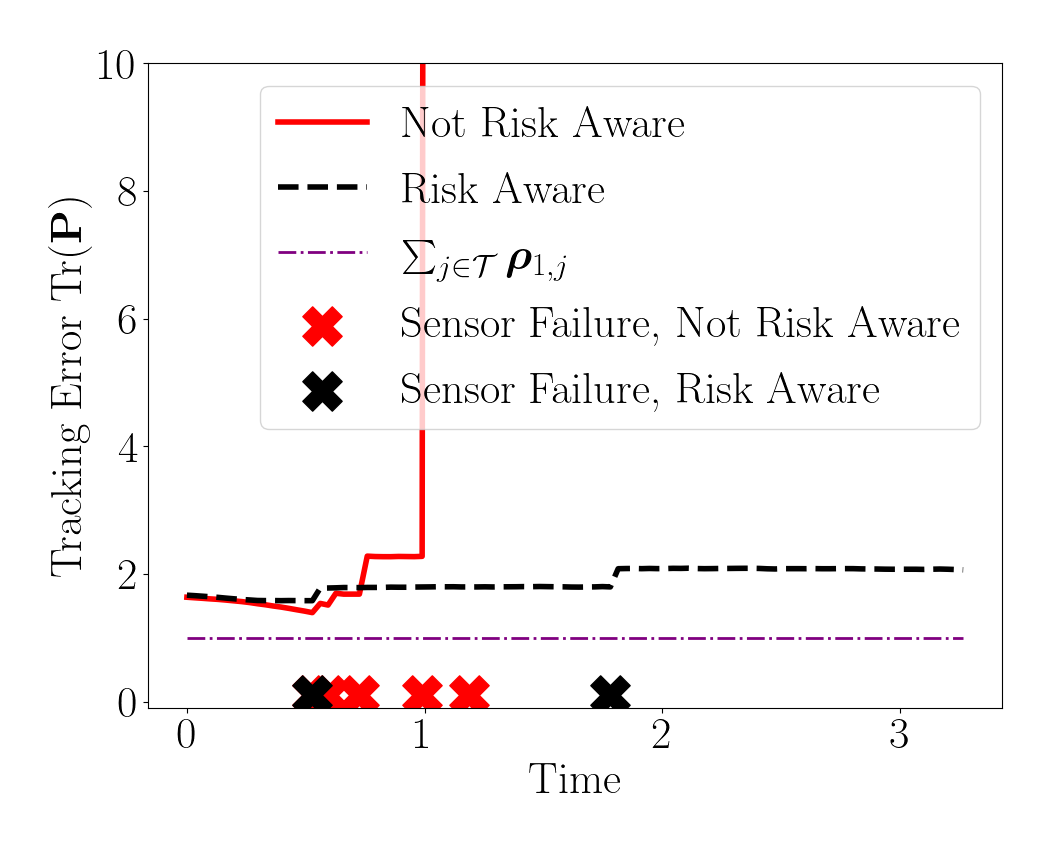}
	\label{fig:perf_comp}
	} \\
	\subfloat[][]{
	\includegraphics[width=0.45\textwidth]{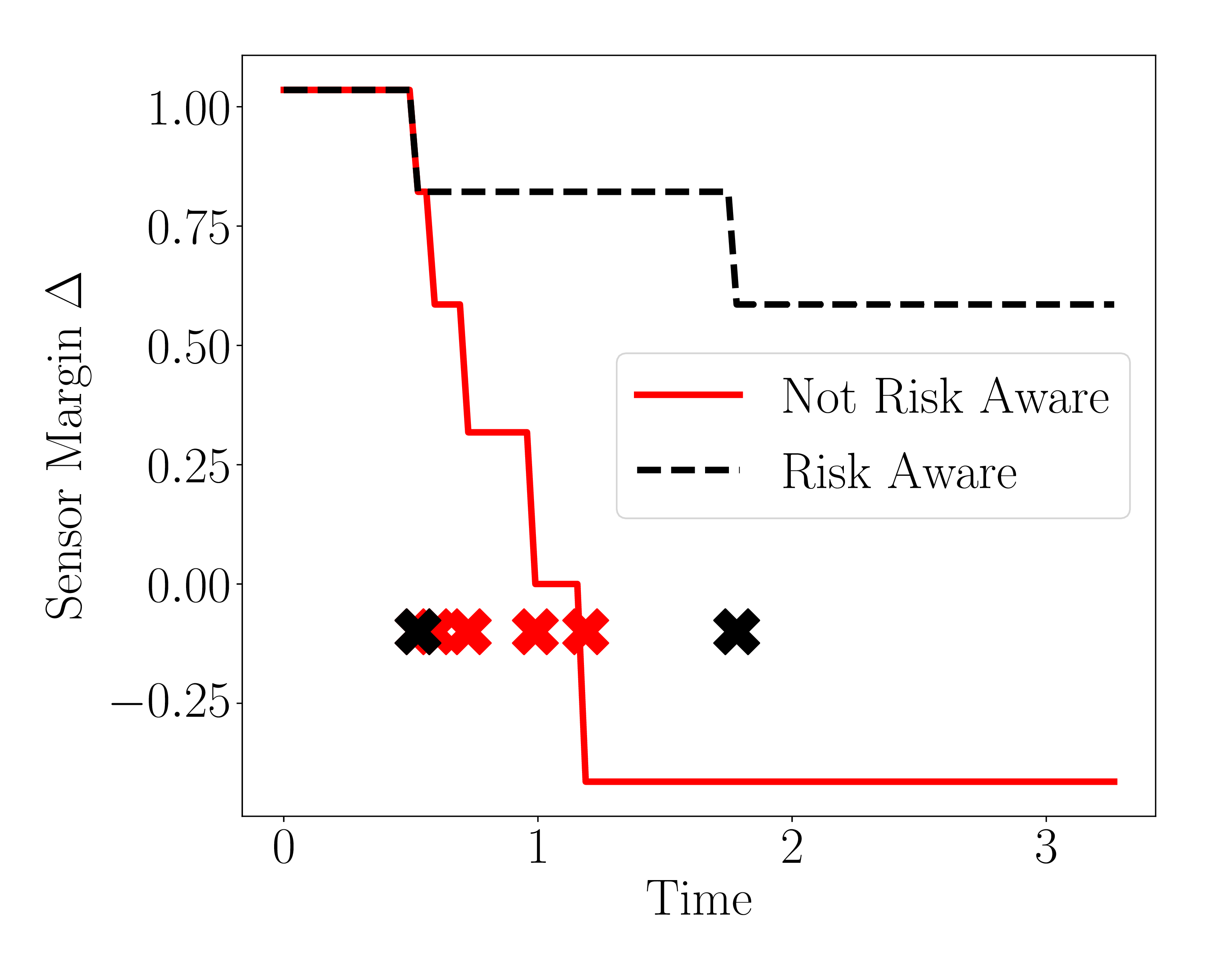}
	\label{fig:sm_comp}
	}
	\caption{Comparison of total tracking error with and without the risk aware controller encoded in constraint~\eqref{eq:nlp:e}. In the case where future performance is not accounted for, the robots drive very close to the target with the aim of meeting their current performance specifications. However, the larger number of sensor failures dramatically increases the tracking error of the robots (see~\protect\subref{fig:perf_comp}). This corresponds to a decreasing sensor margin as seen in \protect\subref{fig:sm_comp}. In contrast, the proposed risk-aware adaptive controller enables the multi-robot team to track the targets for a longer time horizon, and maintains a larger sensing margin $\Delta$ over the simulation time horizon.}
	\label{fig:comp}
\end{figure}

To demonstrate the advantage of the control framework introduced in this paper, we compare the performance of our controller with and without the SOG constraint given by~\eqref{eq:nlp:e}. In this scenario, the robots drive close to the targets in an attempt to obtain accurate estimates of the target positions (see Fig.~\ref{fig:scenario_no_risk}). Figure~\ref{fig:comp} compares the tracking performance and sensor margin over the course of a simulation, and indicates the times at which the sensor failures occurred. As seen, without accounting for the detrimental effects of future failures, the robots drive very close to the targets, and consequently experience a sequence of sensor failures proportional to the target risk field. These simulations demonstrate the ability of our risk-aware adaptive controller to not only continue operations in the face of sensor failures, but also operate for a longer period of time by accounting for future failures. \par

\subsection{Four robots, two targets} \label{subsec:exp2}
\begin{figure}[h]
	\centering
	\subfloat[][]{
	\includegraphics[width=0.425\textwidth]{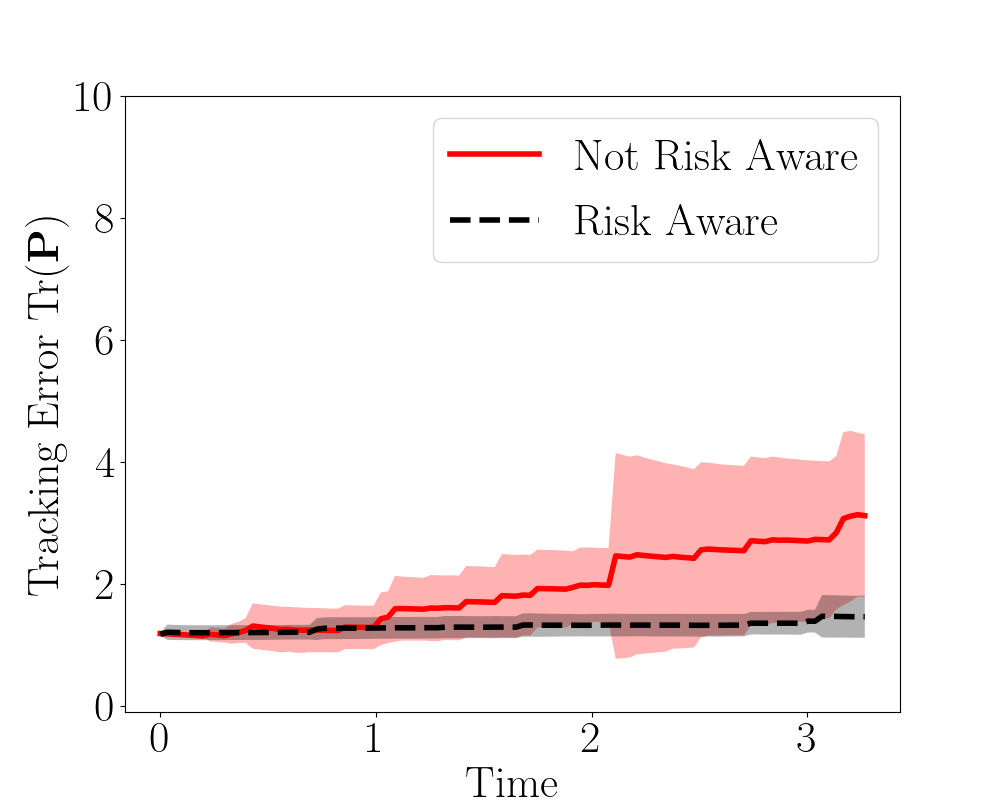}
	\label{fig:ens_perf_comp}
	}\\
    \subfloat[][]{
    \includegraphics[width=0.425\textwidth]{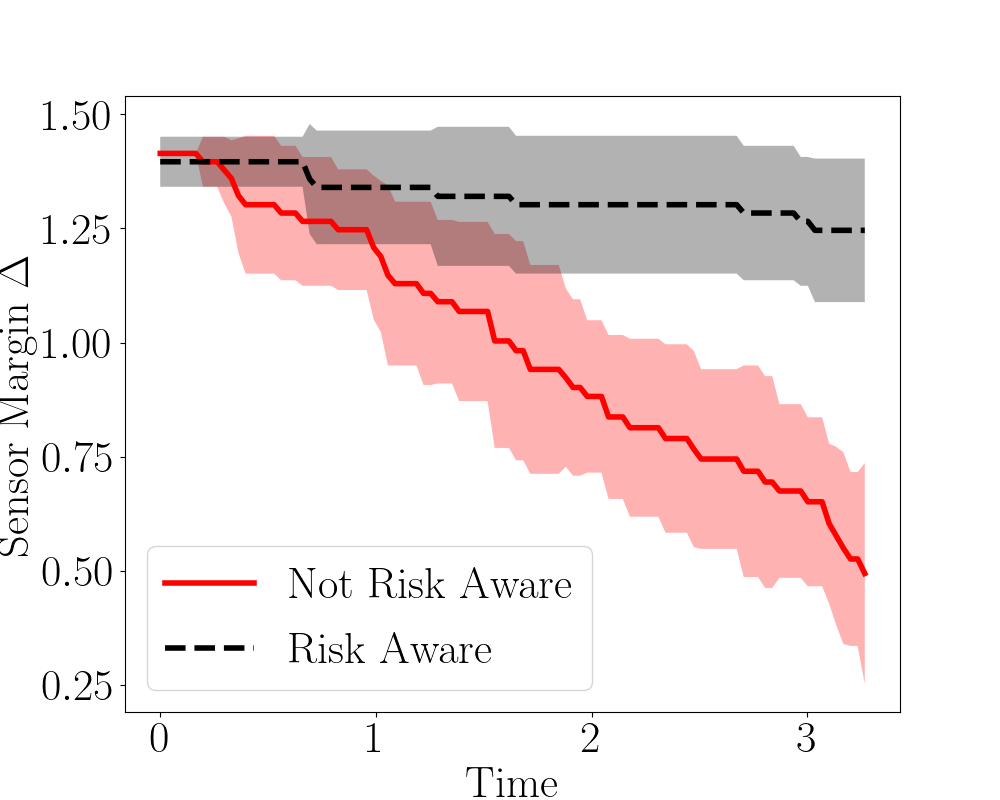}
    \label{fig:ens_sm_comp}
    }		
	\caption{Performance of the proposed adaptive and risk-aware target tracking controller in the face of target-induced failures (4 robots, 2 targets). Similar to Fig.~\ref{fig:comp}, the addition of the risk-aware component encoded by constraint~\eqref{eq:nlp:e} enables the proposed controller to balance performance maximization with the quality of future measurements---allowing for better performance over a longer time horizon. These results were averaged over 10 simulation runs to demonstrate consistent performance. The solid lines denote the mean values, and the shaded region depicts the $\pm 1$ standard deviation. This is especially clear in~\protect\subref{fig:ens_perf_comp} where the variance over the simulation runs is lower in the case of the risk-aware controller.}
	\label{fig:ens_comp}	
\end{figure}

For the second set of experiments, we consider a team of 4 robots, which are equipped with the same types of three sensors described in~\eqref{eq:sensors}. We assign the the following sensor matrix for the team at deployment time: $\bs{\gamma}_1 = [1, 2, 3], \bs{\gamma}_2 = [1, 2, 3], \bs{\gamma}_3 = [3], \bs{\gamma}_4 = [1, 2]$.
For the parameters $(d_m = 0.33, d_n = 2.0, c_j = 1, \Sigma_j = Diag([4.0, 4.0]), \bs{\rho}_1 = [0.45, 0.45], \rho_2 = 0.1, w_1 = 1 , w_2 = 500)$, Fig.~\ref{fig:ens_comp} compares the performance of the algorithm with and without the risk-aware constraint~\eqref{eq:nlp:e}. In particular, we present the averaged results over 10 simulation runs to demonstrate the consistent performance of the developed framework. As seen, not only does the proposed framework ensure the longer operability of the robot team, the variance in the tracking performance is also lower when compared to the case with no risk-awareness. \par 

\begin{figure}[h]
	\includegraphics[trim={0.1cm 0cm 0.0cm 0cm},clip,width=0.47\textwidth]{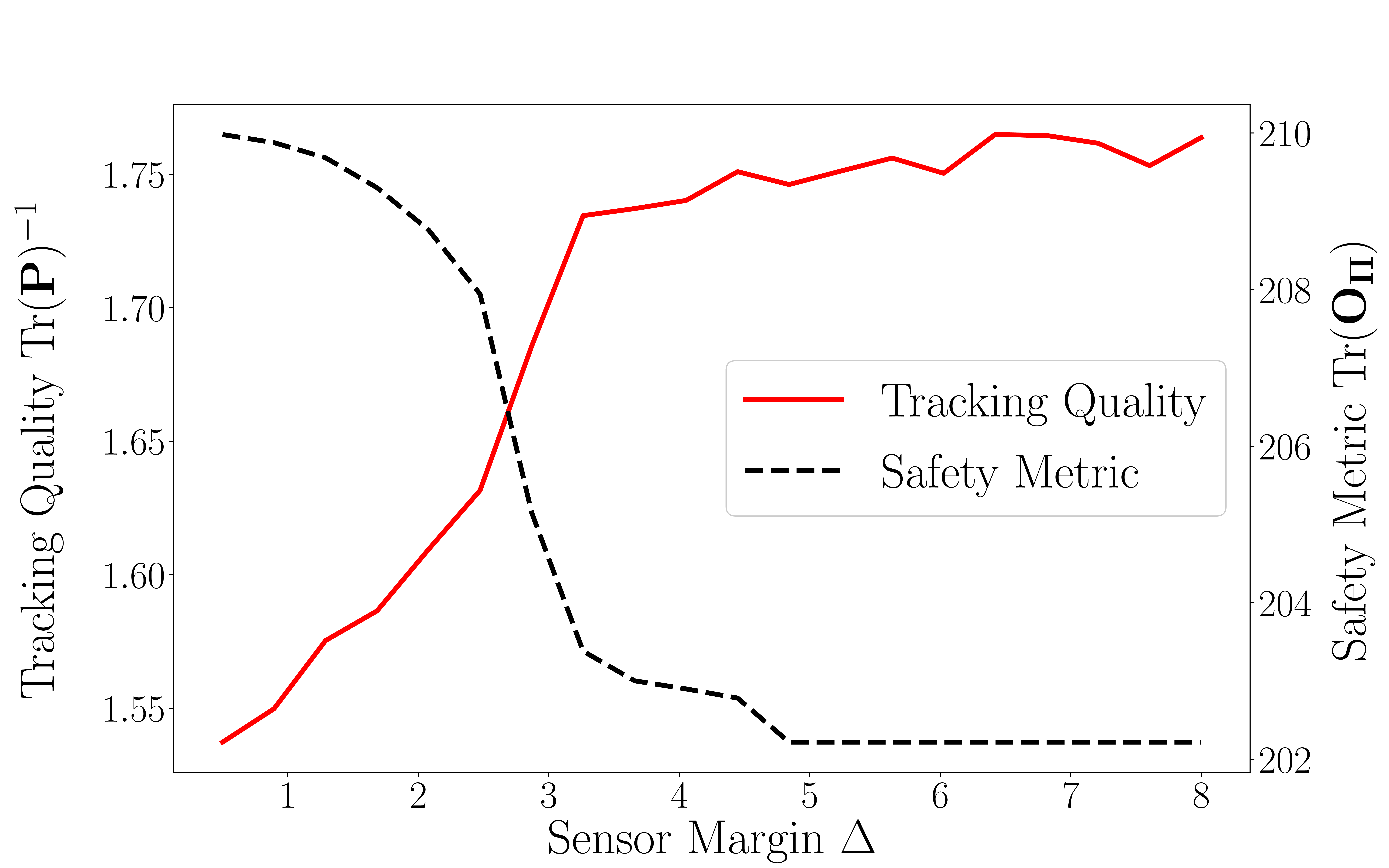}
	\caption{The automatic trade-off between performance and risk-aversion within the proposed control framework as a function of the resource abundance within the team. Thanks to the encoding of sensing margin $\Delta$ within the cost functions $k_1$ and $k_2$ (see~\eqref{eq:hexamples}), the optimization-based controller in~\eqref{eq:full_opt_target_tracking} inherently prioritizes performance for high sensor margins, and prioritizes safety as the sensor margin decreases. This is reflected in the tracking quality and safety metric of the generated robot configurations.}
	\label{fig:sm_tradeoff}
\end{figure}

One of the salient features of the proposed framework is the automatic trade-off between performance maximization and risk-aversion which accounts for the excess of sensors available within the team, as defined in~\eqref{eq:resource_abundance}. Towards this end, Fig.~\ref{fig:sm_tradeoff} compares the team-wide tracking quality $\Tr(\mb{P})^{-1}$ and the safety metric $\Tr(\mb{O}_{\bs{\Pi}})$ for varying sensor margin values. From  Fig.~\ref{fig:sm_tradeoff} it is interesting to observe that the system prioritizes performance over safety when the abundance of sensors in the team is high, but flips this relation as the abundance reduces. Note that, to illustrate this point, the sensor margin $\Delta$ was not computed using~\eqref{eq:resource_abundance} but was directly specified to the optimization program which then generated the robot configurations accordingly. 

\section{Conclusion}\label{sec:conclusion}
This paper developed a framework to leverage the heterogeneous sensors in a robot team to track hostile targets while simultaneously accounting for current and future tracking performance. The cost functions~$k_1$ and~$k_2$, which were a function of the sensor abundance~$\Delta$, are an implicit mechanism to trade-off between risk-aversion and performance maximization. While only a simple version of these functions were considered in this paper, future investigations might reveal a systematic mechanism to design these functions to obtain a desired outcome. 

%% Use plainnat to work nicely with natbib. 

\bibliographystyle{unsrt}
\bibliography{references}

\end{document}